\documentclass[11pt, a4paper, twocolumn]{article}
\usepackage{xurl}
\usepackage[
backend=biber,
style=alphabetic,
minnames=3,
minalphanames=3,
maxbibnames=4,
minbibnames=3, 
maxcitenames=4, 
mincitenames=3
]{biblatex}
\usepackage{abstract}
\usepackage{graphicx}

\usepackage{lipsum}
\usepackage{authblk}
\usepackage{authblk}

\newcommand*{\captionsource}[2]{%
  \caption[{#1}]{%
    #1%
    \textbf{ Source:} #2%
  }%
}

\newcommand{\authinfo}{
    \bigskip
    \begin{flushleft}
    \textbf{Author Information:}\\
    Tomasz Prytu{\l}a\\
    Alexandra Institute\\
    Rued Langgaards Vej 7\\
    2300 K{\o}benhavn S \\
    Email: \texttt{tomasz.prytula@alexandra.dk}
    \end{flushleft}
}

\usepackage{hyperref}
\hypersetup{colorlinks=true, urlcolor=blue, linkcolor=blue, citecolor=blue}

\bibliography{references.bib}

\begin{document}

\title{Graph representations of 3D data for machine learning}

\author{Tomasz Prytu{\l}a }

\newcommand{\abstractText}{\noindent
We give an overview of combinatorial methods to represent 3D data, such as graphs and meshes, from the viewpoint of their amenability to analysis using machine learning algorithms. We highlight pros and cons of various representations and we discuss some methods of generating/switching between the representations. We finally present two concrete applications in life science and industry. Despite its theoretical nature, our discussion is in general motivated by, and biased towards real-world challenges.
}

\twocolumn[
  \begin{@twocolumnfalse}
    \maketitle
    \begin{abstract}
      \abstractText
      \newline
      \newline
    \end{abstract}
  \end{@twocolumnfalse}
]


\section{Introduction}

3D data appears naturally in science and industry, and covers a wide range of domains, from bioimaging (microscopy images, CT scans), through molecular chemistry, to 3D modeling and design plans \cite{3d_everywhere}. A 3D representation is often advantageous as it describes a real world (hence 3D) object more accurately, compared to e.g.,\ 2D projections or slices. However, a drawback of the 3-dimensional representation is the computational cost of the analysis, as the extra dimension, together with scarcity typical for 3D data, makes it very challenging to apply learning algorithms that scale effectively - to the extreme where already analysis of a single sample can be on a verge of capability of a single machine.

In our work we have investigated whether this challenge can be overcome, or at least partially alleviated by employing lighter representations of 3D data - graphs, meshes, point clouds, and simplicial complexes, and the corresponding deep learning algorithms that operate on those representations. Our findings suggest that in many real-world situations it is the case, and we hope that practitioners of machine learning can adapt some of our learnings to their work with 3D data. Our approach is backed by recent developments in the field Geometric Machine Learning, both theoretical \cite{bronstein} and software-oriented \cite{torch_pyg}.

In the specific domain of preclinical research and biomedical imaging, we are plannig to release a set of guidelines for analyzing 3D data using a variety of combinatorial methods, in cases where classical 3D deep learning is not feasible.

Another benefit of using combinatorial representations is their potential for explainability. Since such representations usually come at a higher level of abstraction (e.g.,\ a graph modeling a human pose), their elements (edges, vertices) naturally carry semantic meaning, and thus it may be easier to extract the ‘logic’ behind a machine learning model's predictions.

\section{3D data}

In this section we present an overview of some representations of 3D data, and we compare them from a viewpoint of analysis using deep learning methods. The overview is biased by the specific problems we encountered, and by the overarching theme of studying combinatorial representations.

\subsubsection*{Volumetric}

The main challenge with 3D data compared with 2D data is the computational complexity of algorithms to analyze this data. This stems from the fact the most common representation of 3D data is by voxels (3D analogue of pixels), which means that a volume is represented by a dense grid of 3-dimensional cubes, each of the cubes storing information about e.g.,\ RGB color or intensity. This is a standard format used in many imaging techniques, and thus can be considered as a ‘fundamental’ or ‘raw’ format for 3D data. It is a trivial observation that the number of voxels grows exponentially with the dimension, and thus already a jump from 2D to 3D has severe consequences for compute requirements.

\begin{figure}[!ht]

\includegraphics[]{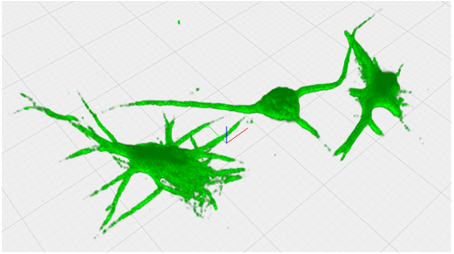}
\captionsource{A 3D image of a neuron cell. The actual cell occupies relatively little of the encompassing 3D volume.}{Bioneer (\href{https://bioneer.dk}{https://bioneer.dk}).}
\label{fig:neuron_cell}
\end{figure}

Another issue, which is somehow more a characteristic of 3D data in general, is its sparsity. Whether it is a microscopy image of a neuron cell, or a 3D design for manufacturing, a lot of voxels are unoccupied, and the actual object of interest fills only a small portion of the volume, see Figure~\ref{fig:neuron_cell}. However, one does not know it in advance, and therefore the standard 3D algorithms (like e.g.,\ Convolutional Neural Networks) process the entire volume voxel by voxel.

Finally, 3D data is typically more scarce than 2D data: in many cases one uses an expensive and time consuming procedure to produce a single 3D image (e.g.\ 3D microscopy) and thus by default one deals with much smaller datasets than in other contexts.

An important feature of volumetric representations it that it is a so-called \begin{it}Euclidean representation\end{it}, meaning that there is a global coordinate system of XYZ coordinates. Thus, after agreeing on a given size of the volume, all the examples have the exact same size, and thus the algorithm for analysis does not have to accommodate size differences between samples.

There are a plenty of other ways to represent the 3D data (and switch between the representations) that can have some advantages. All the following representations are the examples of non-Euclidean representations.

\begin{figure*}[!ht]

\includegraphics[width=\textwidth]{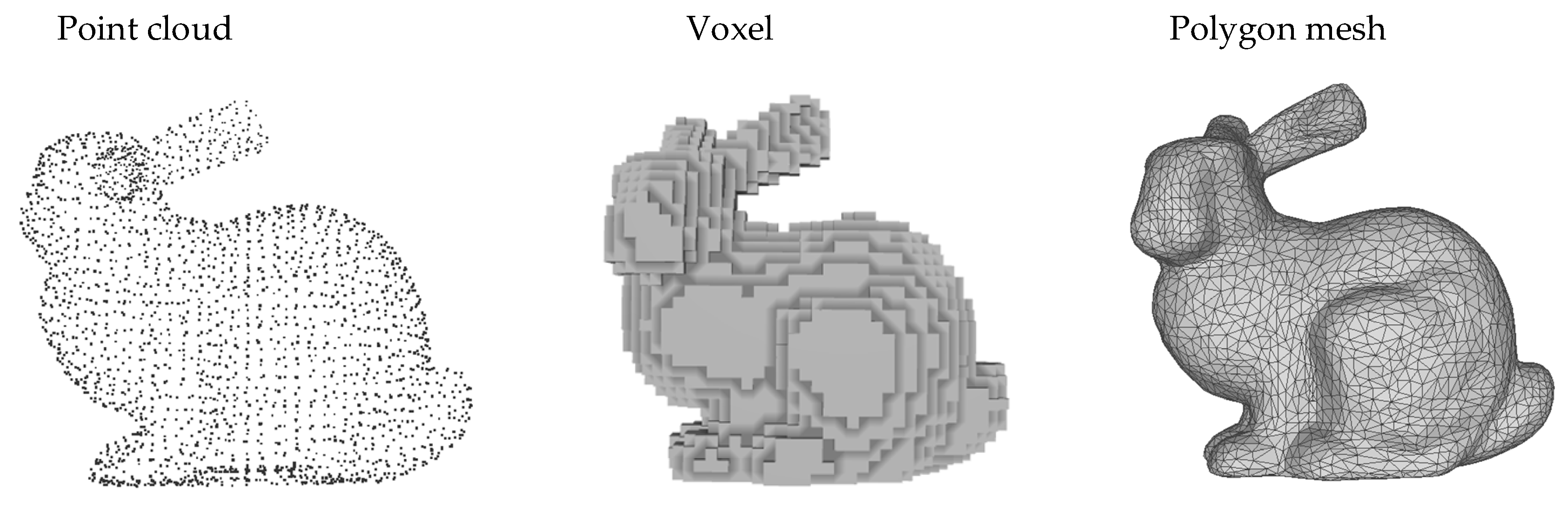}
\captionsource{Example of various data representations of a 3D object.}{\cite{rabbit_fig}.}
 \end{figure*}

\subsubsection*{Mesh}

Mesh is a tessellation of a surface of a 3D object by triangles (or other polygons). It describes geometry well and removes the sparsity problem. This is because to represent a mesh, one only needs to store the position of vertices and encode which vertices form triangles/polygons. As such, the mesh format does not offer an intrinsic global coordinate system, but it has a resemblance of the coordinate system locally, as the neighborhoods of vertices are all disks. This means that around every vertex, mesh looks like a planar region, and thus one can use techniques for processing 2D images, appropriately adjusted. From the deep learning viewpoint there are a few architectures to handle meshes (mostly variants of MeshCNN \cite{mesh_cnn}), which cleverly exploit geometric features of the mesh without referring to its embedding into $\mathrm{R}^3$ (like angles and edge lengths). There are also some novel pooling and unpooling operations which are specific to mesh format.

However, by definition a mesh comes in quite a constrained form, i.e.,\ it is a surface of a 3D manifold. This is a strong restriction on topology, which is great for visualization/rendering, but from the point of view of analysis it is often not flexible enough. Especially when the data is highly non-smooth or non-manifold. As shown in Section~\ref{subsec:rd8}, even if data at hand is smooth, the mesh format does not allow easy augmentation or transformation of the data.

\subsubsection*{Point cloud}

A point cloud is a collection of points in a 3D space, described by their coordinates together with some additional features, like RGB color or the intensity value. They ‘occur naturally’, e.g.,\ from Lidar scanning. Point cloud offers a lot of flexibility of non-Euclidean data, together with some structure of Euclidean data. However, often point cloud format is too loose to capture geometric information - as there is no apparent relation between the points, besides what one can infer from their attributes. For example, given only points and no other information like edges or triangles, it may be impossible to tell two objects apart.
Another challenge is the size - point clouds often contain millions or even billions of points. This, however, can be remedied, as it is relatively easy to subsample points. This in fact offers some extra flexibility, as one can choose different granularity for different parts of the object, based on the level of detail one wishes to capture.

\subsubsection*{Graphs}

Many tasks in 3D analysis require understanding the geometry of the object (e.g.,\ structure of the cells and their connections, shape of a molecule, spatial relations, proximity of the parts in 3D designs). Keeping that in mind, graphs as a data representation seem to be a good balance between expressiveness and simplicity. Graphs can be defined in a purely abstract way - one needs to specify the set of vertices and tell which vertices are connected by the edges, yet they often naturally carry spatial and geometric information, when one pictures a graph as a network of nodes in 3D (from that viewpoint one can think of graphs as ‘enhanced’ point clouds, where one pre-specifies which nodes are more ‘important’ to each other by joining them by edges).

There are many flavors of graphs seen as data representation. One particularly useful for lowering the dimension is \emph{skeletal graphs}, that is, graphs which serve as a $1$-dimensional skeleta for complex higher dimensional geometric objects \cite{skeletons}. An example could be a schematic graph representing human body, with nodes corresponding to joints, and edges corresponding to legs, hands, torso, etc.\ Another flavor is knowledge graphs, where nodes correspond to entities, and edges describe various relations between them, e.g.,\ social networks. These types of graphs though have in principle nothing geometric about them. One also finds graphs which are schematic representations of actual object, e.g.,\ molecules in chemistry. 

From the machine learning viewpoint, graphs come together with a custom range of architectures to analyze them, the so-called graph neural networks (GNN’s). Graphs have plenty of unique advantages over other data formats:
\begin{itemize}
\item{\textbf{Scalability:}} due to sparsity of 3D data, when moving from 2D to 3D one does not observe a big change in the number of vertices or the size of the features (for example,  a positional encoding simply gets an extra column of z-coordinates). The GNNs for 3D  graph data can proceed almost as fast as for 2D data, as they simply do not see the dimension but only the connections between the nodes. This is in high contrast with voxels, where both the size of the data, and the training time gets significantly higher when moving to 3D. (On the other hand, for processing a regular grid of 2D points, like e.g.,\ an image, a CNN will be faster than a GNN).

\item{\textbf{Ease of engineering and adding features:}} it is customary to have a graph together with a vector of vertex attributes and edge attributes, for each vertex and each edge respectively. The common practice is to add various either externally known information (e.g.,\ type of vertices if they represent different entities, their color, etc.), or information stemming from a graph itself and its positional encoding (e.g.,\ degree of vertices, lengths of edges, angles between the edges, etc.). In principle one can do similar things for voxels, it feels though that lack of naturally occurring features prevents that in practice.

\item{\textbf{Flexibility in enriching or transforming the graph itself:}} one can easily add extra edges and vertices, to encode vital information e.g.,\ join vertices by edges based on their proximity wrt. a chosen metric, or add ‘higher order’ nodes which group all vertices of a given type. Note that this is not possible for e.g.,\ meshes, as adding an edge without a triangle, will cause it to stop being a mesh.

\end{itemize}
\subsection{The ‘2-step process’ in graph analysis}

Given a 3D data representation, to benefit from Graph Neural Networks and other graph related tools, one first must turn the data into a graph format. This is the first step of the 2-step process, the second step being the analysis of the graph itself. This task can already be nontrivial, where one may need to use domain specific knowledge and various tricks. There are a couple of standard methods, but they usually do not lead to optimal graphs:
\begin{itemize}
\item{Mesh $\rightarrow$ graph:} take the 1-skeleton, i.e., only vertices and edges, forget about triangles/polygons. (Loses face information. One can subdivide mesh beforehand to retain some info about polygons.)
\item{Voxel $\rightarrow$ graph:} let voxels be nodes and connect every voxel to 6 of its neighbors. (Results in a graph as large as the volume itself.)
\item{Point cloud $\rightarrow$ graph:} add edges based on the proximity of vertices (may result in a very large number of edges, if the distance threshold is too low). Note that ‘proximity’ here does not necessarily mean spacial proximity, but can be an arbitrary user-chosen metric.
\end{itemize}
It is worth mentioning that some architectures, e.g.,\ PointNet++ \cite{pointnet_pp} create the graph dynamically while training for a downstream task, thus effectively merging steps 1 and 2 into one pipeline.

\subsubsection*{Graph generation methods}

We investigated mostly the so-called ‘skeletonization methods’, whose aim is to produce a $1$-dimensional skeleton of a 3D geometric shape, which captures global geometric information to a good degree of detail. These include classical methods such as signed distance function thinning, or Medial Axis Transform as well as machine learning-inspired approaches, which focus on creating a graph that is amenable directly to analysis with GNN’s, e.g.,\ SN-graph \cite{sn_graph}. 

\begin{figure}[!ht]

\includegraphics{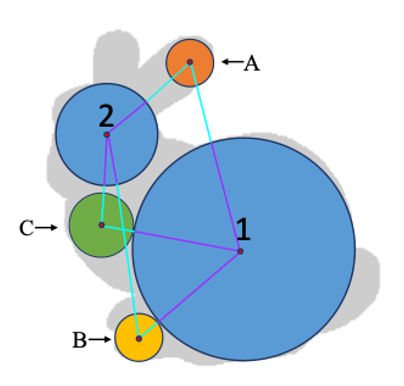}
\captionsource{The skeletonization algorithm choses nodes based on the size of the sphere one can inscribe in the object.}{\cite{sn_graph}.}
 \end{figure}

\begin{figure*}

\includegraphics{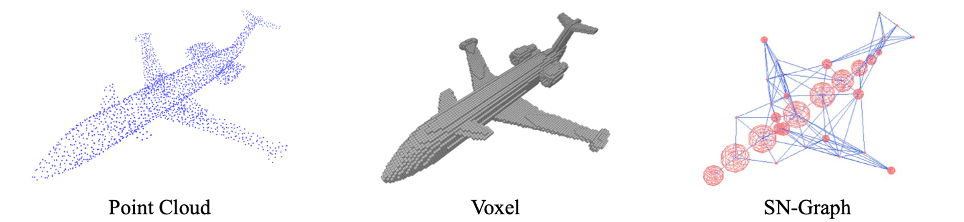}
\captionsource{Graph skeleton (right) is considerably lighter than raw voxels (center) or dense point cloud (left).}{\cite{sn_graph}.}
 \end{figure*}

\subsubsection*{Analysis methods}

A default choice for graph data is that of Graph Neural Network (GNN). The key characteristic of these architectures is how to apply ‘filters’ on a vertex level and how to aggregate information from neighbors to update the state of the vertex (the so-called \emph{message passing}). In that regard, Graph Neural Networks can be seen as analogues (or even a generalization) of Convolutional Neural Networks to graph inputs. 

There is also some novelty in how to perform graph pooling and unpooling operations, and how to perform e.g.,\ dropout, but many other standard neural network components can be incorporated without much trouble (e.g.,\ attention mechanism, batch normalization, etc.). This gives a lot of flexibility in designing graph networks, especially given that there exists a Pytorch-based library at one’s disposal \cite{torch_pyg}.

One can also resort to a simpler approach of vertex embeddings (e.g.,\ node2vec \cite{node2vec}), to map a graph back to the Euclidean space, where more standard architectures are available.

Another ‘hack’ that can sometimes be beneficial is to use the graph representation as a source of features for classical models, like e.g.,\ Decision Trees. This approach works if a graph is supposed to describe some quantitative features of the object of interest.

\subsubsection*{Graph analysis vs. 3D CNN analysis} Since to train a GNN one has to first convert the data into a graph, then in order to compare computational complexity of this approach, to e.g.,\ training a 3D CNN directly on volumetric data, one has to take into account the complexity of both the conversion step and then the GNN pipeline. While we indicated that in practical applications graph neural networks will be faster (mostly because the input graph is smaller than the original volume) it is still to be discussed how costly the conversion process is. Many of the skeletonization algorithms are not learning algorithms, but rely on hard-coded logic, and thus are relatively fast to apply (even when testing a couple of configurations of hyperparameters). But somehow more importantly, it is the deep learning part of the analysis that is extra costly, and this is because in practice one trains not one, but several models with some changes of architecture and hyperparameters along the way, before one arrives at a satisfying result. So, in the long run, having a more compact representation will always be beneficial.

\section{Applications in concrete cases}

In this section we present findings from two projects in collaboration with life scientists and a software company, respectively.

\subsection{Mitochondrial networks in muscle cells}

The area of bioimaging and preclinical research and development is a particularly good testbed for our programme, as one almost always deals with the major challenges of 3D data: the large sample size, sparsity, and scarcity. At the same time there is a lot of variation across different tasks, so that almost every single imaging problem requires a tailored pipeline for the analysis. We present one such case, which is a joint work with Clara Prats \cite{mito}.

\begin{figure}[!ht]

\includegraphics[width=\columnwidth]{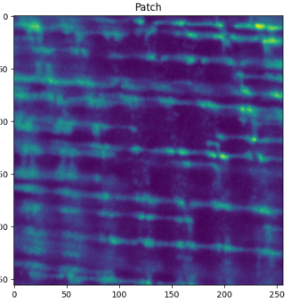}
\caption{A slice of an image of a mitochondrial network.}
\label{fig:slice}
\end{figure}

\subsubsection*{Data} Data comes as confocal fluorescent microscopy images of muscle cells, multichannel, with one channel showing mitochondrial networks which are crucial for the task.  Manually labelled on a scale from 1 to 10, by an expert annotator.

\subsubsection*{Task} Predict the degree of healthiness of mitochondrial networks. It can be posed as a binary classification healthy/unhealthy (healthy: score larger than 5, unhealthy: score at most 5), or a regression on a scale 1 to 10.  Healthiness of these networks is associated with several conditions, such as PCD (primary carnitine deficiency, a certain muscle disorder) \cite{mito_msc}, obesity, physical inactivity, and type 2 diabetes \cite{obesity_mito}. 

\subsubsection*{Data description} A mitochondrial network resembles a rectangular grid, see Figure~\ref{fig:slice}. The images contain considerable amount of noise and some imaging artifacts. The criteria behind annotation were: the more healthy the cell is, the more regular and straight the grid is, the fewer bursts or ruptures one can find in it.

\subsubsection*{Method}
Given the intuition behind the healthiness of cells we thought of a graph as a natural representation to encode the images, as it precisely captures connectedness of the grid and its geometric shape. Due to the shallowness of volumes (up to $20$ voxels in z-direction) we focused on testing the approach on 2D slices. However, the approach will scale with no problems to deeper volumes in 3D. Furthermore, to reduce computational complexity, we cut every such slice into patches of size $256 \times 256$ pixels each - the original slices have varying sizes of up to $1500$ pixels in x- and y-direction. Also, having many 'small' samples as opposed to few 'big' ones can help in convergence and stability of training of the machine learning model.

\subsubsection*{Graph generation} 
To create graphs, we use a variant of skeletonization algorithm called the \begin{it}SN-graph\end{it} \cite{sn_graph}. We introduced some minor tweaks to the algorithm to obtain graphs that are visually closer to the images. The tuning of hyperparameters for the algorithm was performed based on expert intuition and to some degree guided by model's performance - generation of the graph is a deterministic, yet costly process, so only a few configurations of parameters were tested. 

Dimensionality reduction: The patch in Figure~\ref{fig:slice} has $256 \times 256$ pixels. We constrained graphs to have maximally $300$ vertices, and on average such a constraint resulted in around $600$ edges (on a perfectly rectangular grid, one has asymptotically 2 times as many edges as vertices, so this result gives us an implicit validation of the graph generation algorithm).

\begin{figure*}[!ht]

\includegraphics{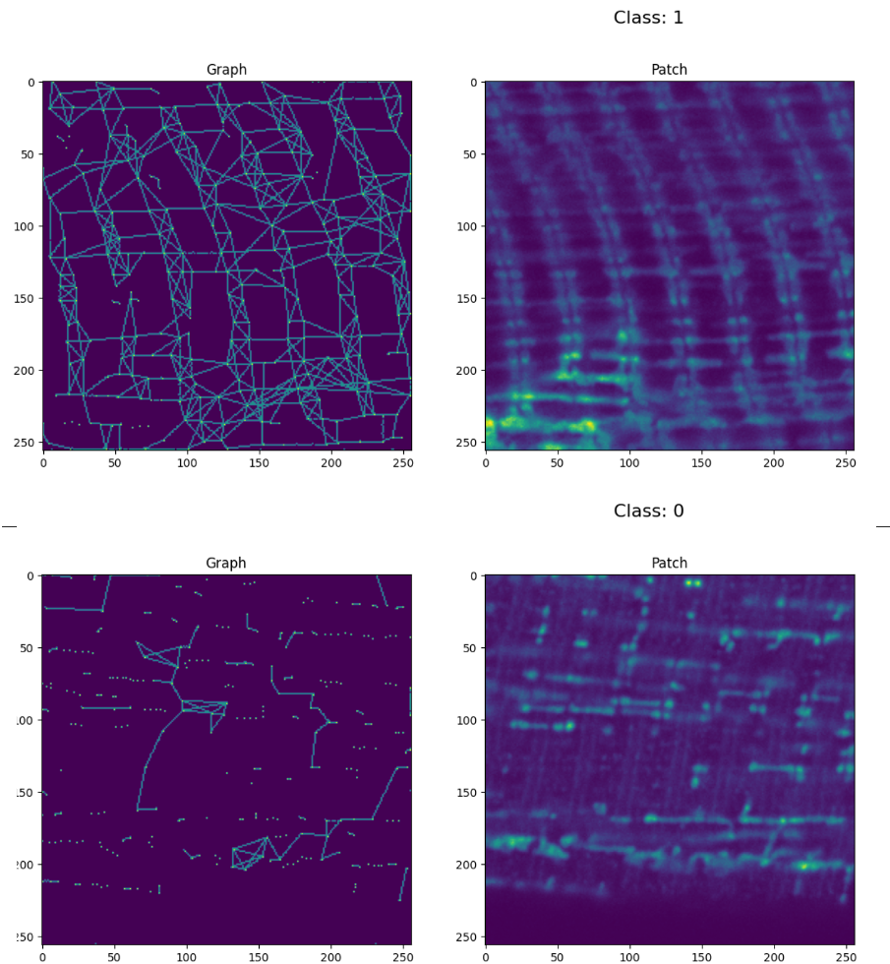}
\caption{Graphs generated for images for two classes. For image classified as 1=healthy the corresponding graph (upper left) is fairly regular. The graph corresponding to class 0=unhealthy (lower left) is degenerated.}
\end{figure*}

\subsubsection*{Classification/regression model} 

We use Graph Attention Network \cite{gat}, with 4 attention layers, and 2 fully connected layers. The vertex and edge attributes include standard geometric features, such as lengths of edges, their slope (seen as a discrete variable: vertical, horizontal, or skew), degree of vertices, etc. We also include some features of an underlying image in the neighborhood of a vertex, e.g.,\ the average intensity. For a comparison we also use a convolutional neural network for the same task.

By aggregating prediction over all slices for each volume, both approaches resulted in $100{\%}$ accuracy on the binary classification task. On the single ‘patch’ level, GNN achieved $0.86$ accuracy, whereas CNN achieved $0.95$ accuracy. On the flipside, GNN was trained on a single CPU, in a little less than an hour, while CNN required a GPU acceleration.

Having generated enough quantitative, global features such as e.g.,\ the average degree of a graph, the number of horizontal/vertical edges, the average intensity of vertices, etc., we tried to make a classification based solely on those features with a use of the Random Forest algorithm. Surprisingly, we also got $100{\%}$ accuracy on the volume level, and $0.84$ on patch level. On the other hand, these features intuitively do describe how ’regular’ the graph is, so maybe such a high performance of a model based on those features is not that surprising. But nonetheless, we believe that in general skeletonization algorithms can be thought as good feature extractors, regardless of the final model used for the task.

\subsection{AI for 3D design generation and analysis}\label{subsec:rd8}

\begin{figure}[!ht]

\includegraphics[width=\columnwidth]{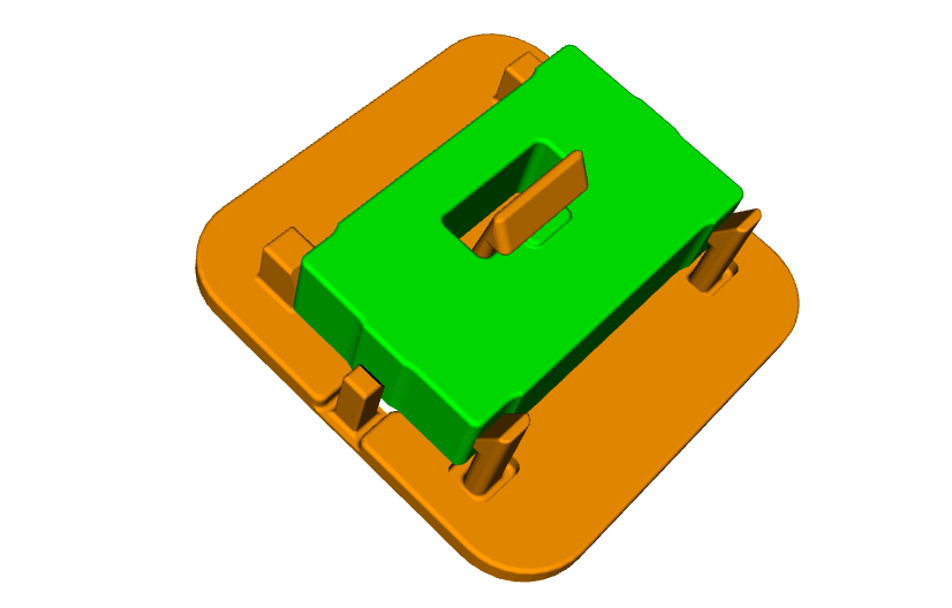}
\caption{A pair of interconnected parts from a 3D CAD drawing.}
\label{fig:cad_drawing}
\end{figure}

We present findings from a pilot project under the AI:Denmark initiative (\href{https://aidenmark.dk/}{https://aidenmark.dk/}) with a software company called RD8 Technology (\href{https://rd8.tech}{https://rd8.tech}). RD8 Technology investigates 3D design drawings for manufacturing, and offer improvements to those designs, from the point of view of functionality, material use, durability, robustness to manufacturing imperfections, etc.\ They work with 3D CAD models (see Figure~\ref{fig:cad_drawing}) and process them using, among other measures, parametric and analytic geometry. They plan to introduce AI and a data-driven approach to improve their model application and functionality. There are multiple possibilities for doing so, to name a few: an AI-aided automatic generation of designs, automatic suggestions for improvements to existing designs, or detection of bottlenecks or errors. Behind many of these improvements there is a need for an ML-architecture capable of learning geometry, topology, and all the physical constraints (like e.g.,\ movement or bending of parts). In our pilot project, we isolate a fairly simple task to see how one can begin creating such an architecture. 

\subsubsection*{Data} Data comes as parametric 3D geometries, and as such it is almost hopeless to try to analyze it using 3D convolutional neural networks. This is because a generic design plan contains details on many different scales and thus, to make a putative voxel representation one needs to use very high resolution, which could result in a gigantic dimension (in terms of the amount of voxels) and a lot of superfluous information. Even more importantly, the parametric description is at a much higher level of abstraction than voxels, therefore turning it into volume seems like an unnecessary complication. 
On the other hand, a CAD plan can be turned to e.g.,\ polygonal mesh with almost no loss of information. In general, meshes and graphs are naturally very well suited for storing data with varying resolution and amount of details.

\subsubsection*{Task} 
\begin{figure*}

\includegraphics{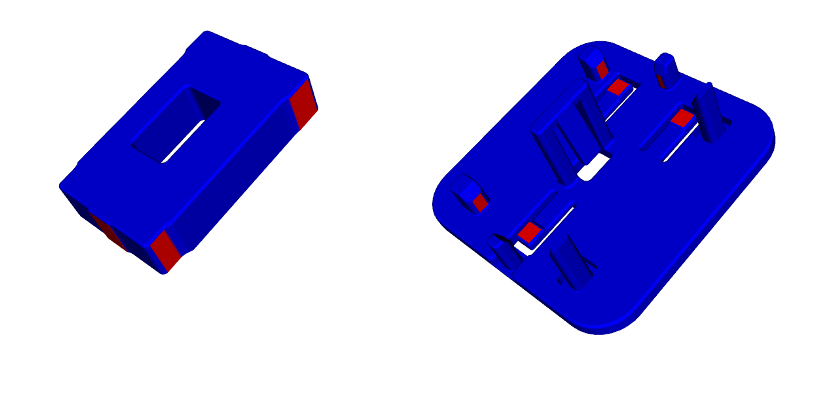}
\caption{A pair of parts, separated. Contact areas are red.}
\label{fig:contacts_separated}
\end{figure*}

A simple test task is the following: given a pair of parts that are interconnected, calculate their exact \begin{it}contact areas\end{it} - areas where the surfaces of the two parts will touch each other, see Figure~\ref{fig:contacts_separated}. More generally, to accommodate practical constraints such as manufacturing imperfections or to determine tolerance for errors, one is also interested in predicting the prospective contact areas if parts nearly touch, or the overlap areas if parts overlap. 

While this task can be solved with linear algebra and analytic geometry, it is interesting to see how close one can get with data-driven methods. There is also a chance for actual improvement. Since the current pipeline is computationally involved and includes many tedious corner cases requiring extra hard-coded rules to perform well, one could hope that a robust learning algorithm could perform the task faster and better capture non-standard patterns.

\subsubsection*{Method} 
\begin{figure*}

\includegraphics[width=\textwidth]{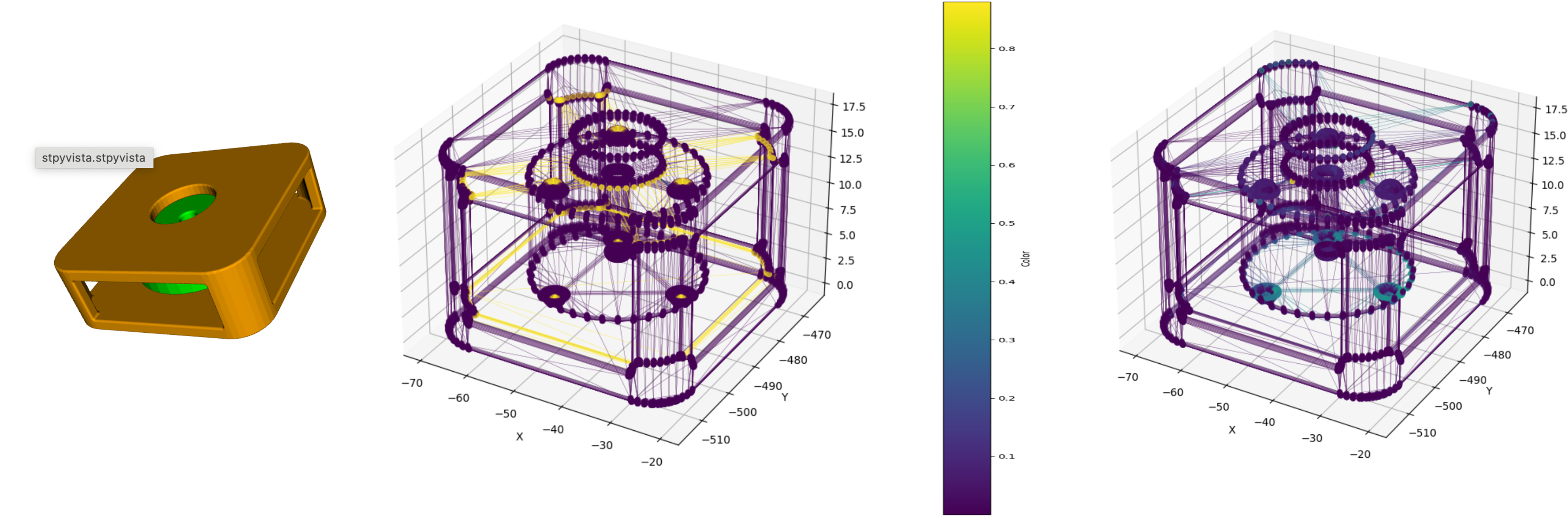}
\caption{Left: Mesh rendering of a pair of geometries. Center: Graph representation, ground truth vertex coloring.  Right: Graph representation, model's prediction coloring. The model assigns higher prediction values to the actual contact vertices.}
\label{fig:visual_inspection}

\end{figure*}

We begin by representing both parts as triangular meshes. In order to enable a broader range of architectures and to be able to incorporate additional information like e.g.,\ spatial relation between the two parts, we turn the meshes into graphs, by ‘forgetting’ the triangles. That is, we remember vertices and which of them are connected by edges, and simply forget the information of which triples of edges/vertices form triangles, see Figure~\ref{fig:visual_inspection}. This loses very little information, as in general there are very few triples of vertices which are pairwise connected but do not span a triangle.

Given such a graph input, the task becomes a binary semantic segmentation of vertices: given a vertex of either of the two parts, a model has to predict whether it is a part of a contact area or not, based on a target mask provided by RD8’s current software (with manual intervention if needed).

Having a graph representation has many advantages. First, since the task depends only on the relative position of two parts, and not on their embedding into the Euclidean space, the most effective choice is a representation that is invariant to the rotations and translations, and this is easiest achieved with a graph or a mesh.
Second, a graph allows us to easily augment our input in many ways:
\begin{itemize}
\item Since the proximity between the parts is a vital feature for the task, we add edges between vertices belonging to different parts which are close to each other. 
\item Because the concept of orientation matters (i.e.,\ what is inside and what is outside of the surface), adding vertex normal vectors seems like a good idea. To do it in a coordinate-free way, one can e.g.,\ add a new vertex at the tip of a normal vector and connect it by an edge to the original vertex.
\end{itemize}

Note that while the notions of ‘proximity’ and ‘orientation’ are somehow available at hand in volumetric representation, the lack of rotational and translation invariance would likely require much more data until the model learns the actual geometric features. On the other hand, as shown above, it is relatively easy to incorporate these features in the graph representation, with a minor increase in complexity.
\begin{figure}

\includegraphics[width=\columnwidth]{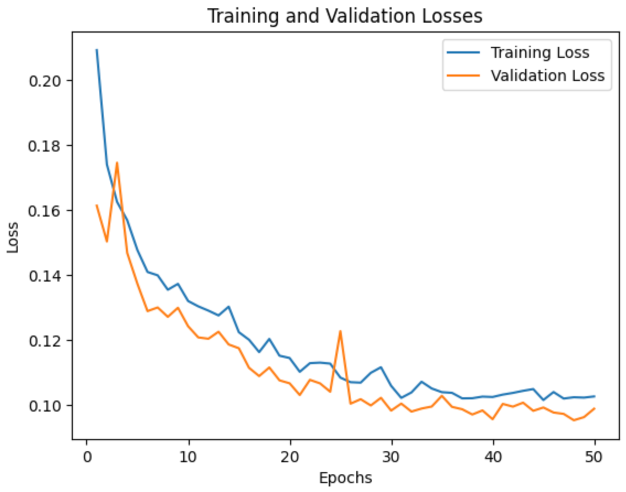}

\caption{Training and validation losses for semantic segmentation on vertices.}
\label{fig:losses}
\end{figure}

The neural network we use is a Graph Attention Network \cite{gat}, with 3 message-passing layers and 2 fully connected layers. The output of the network is a prediction between 0 and 1 for each vertex, of how likely this vertex is to be a part of a contact area.

\begin{figure*}

\includegraphics[width=\textwidth]{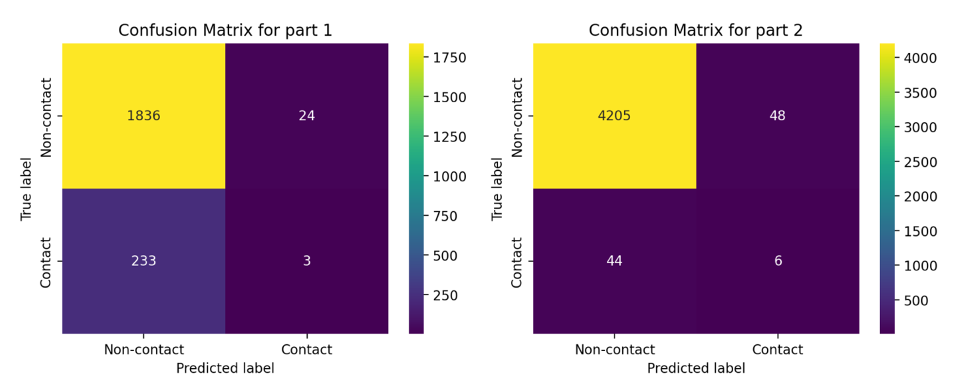}
\caption{Confusion matrices for binary classification of vertices for a pair of parts.}
\label{fig:cm}

\end{figure*}

\subsubsection*{Results} Due to the small amount of training data, we only got an indication that the chosen data representation and the corresponding architecture can learn the task. Namely, as shown in Figure~\ref{fig:losses}, the model trains and generalizes. However, confusion matrices for binary classification presented in in Figure~\ref{fig:cm} show that the model overfits to non-contact vertices (as there are many more of them in each pair of parts). On the other hand, a visual inspection of the predicted areas shows that the model begins to assign higher prediction values for contact areas, even though they are still below the classification threshold, see Figure~\ref{fig:visual_inspection}. This can be taken as a strong indication that with a larger dataset (and possibly a larger network) the model can become quite accurate, and suggests that the GNN architecture is capable of learning the geometry behind the task.


\nocite{*}
\printbibliography

\authinfo

\end{document}